\documentclass[10pt,twocolumn,letterpaper]{article}

\usepackage{cvpr}              

\usepackage{graphicx}
\usepackage{amsmath}
\usepackage{amssymb}
\usepackage{booktabs}

\usepackage[utf8]{inputenc} 
\usepackage[T1]{fontenc}    
\usepackage{url}            
\usepackage{booktabs}       
\usepackage{amsfonts}       
\usepackage{nicefrac}       
\usepackage{microtype}      

\usepackage{multirow}
\usepackage{xspace}
\usepackage{mathtools}
\usepackage{wasysym}
\usepackage{marvosym}
\usepackage{xcolor}
\usepackage{color}
\usepackage{tabularx}
\usepackage{float}
\usepackage{diagbox,tabu,stackengine}
\usepackage{bbm}
\usepackage{bm}
\usepackage{mwe}
\usepackage{graphbox}
\usepackage{sidecap}
\usepackage{gensymb}

\usepackage{algorithm}
\usepackage[noend]{algpseudocode}
\usepackage{comment}
\usepackage{cancel}
\usepackage{ulem}
\usepackage{array}
\newcommand{\PreserveBackslash}[1]{\let\temp=\\#1\let\\=\temp}
\newcolumntype{C}[1]{>{\PreserveBackslash\centering}p{#1}}
\newcolumntype{R}[1]{>{\PreserveBackslash\raggedleft}p{#1}}
\newcolumntype{L}[1]{>{\PreserveBackslash\raggedright}p{#1}}

\usepackage{xspace}
\makeatletter
\DeclareRobustCommand\onedot{\futurelet\@let@token\@onedot}
\def\@onedot{\ifx\@let@token.\else.\null\fi\xspace}
 
\def\ie{\textit{i.e}\onedot}

\def\etal{\textit{et al}\onedot}
\makeatother

\definecolor{primary_pink}{rgb}{0.8, 0.508, 0.467}
\definecolor{secondary_blue}{rgb}{0.32, 0.624, 0.8}
\definecolor{mark}{rgb}{0.97, 0.73, 0.0}

\newcommand{\ourmodel}{\textsc{FEGR}}
\usepackage[pagebackref,breaklinks,colorlinks]{hyperref}

\usepackage[capitalize]{cleveref}
\usepackage{enumitem}
\crefname{section}{Sec.}{Secs.}
\Crefname{section}{Section}{Sections}
\Crefname{table}{Table}{Tables}
\crefname{table}{Tab.}{Tabs.}

\def\cvprPaperID{8120} 
\def\confName{CVPR}
\def\confYear{2023}

\newcommand{\image}{\mathbf{I}}
\newcommand{\camera}{\mathbf{c}}

\newcommand{\density}{\rho}

\newcommand{\mesh}{\mathcal{S}}
\newcommand{\normal}{\mathbf{n}}

\newcommand{\albedo}{\mathbf{k}_d}
\newcommand{\light}{L}
\newcommand{\lightdir}{\bm{\omega}}
\newcommand{\dir}{\mathbf{d}}
\newcommand{\brdf}{f_r}
\newcommand{\visibility}{v}

\newcommand{\origin}{\mathbf{o}}
\newcommand{\ray}{\mathbf{r}}

\newcommand{\normalmap}{\mathcal{N}}
\newcommand{\materialmap}{\mathcal{M}}
\newcommand{\albedomap}{\mathcal{K}_d}
\newcommand{\depthmap}{\mathcal{D}}

\begin{document}

\title{Neural Fields meet Explicit Geometric Representations\\for Inverse Rendering of Urban Scenes}

\author{
  Zian Wang$^{1,2,3}$ 
  \quad Tianchang Shen$^{1,2,3}$ 
  \quad Jun Gao$^{1,2,3}$ 
  \quad Shengyu Huang$^{1,4}$ 
  \quad Jacob Munkberg$^{1}$ \\
  \quad Jon Hasselgren$^{1}$ 
  \quad Zan Gojcic$^{1}$ 
  \quad Wenzheng Chen$^{1,2,3}$ 
  \quad Sanja Fidler$^{1,2,3}$ \\
  $^1$NVIDIA \quad $^2$University of Toronto \quad $^3$Vector Institute \quad $^4$ETH Z\"{u}rich \\
}

\twocolumn[{
\renewcommand\twocolumn[1][]{#1}%
\maketitle
\begin{center}
\vspace{-3mm}
\includegraphics[width=0.95\linewidth,trim=0 0 0 0,clip]{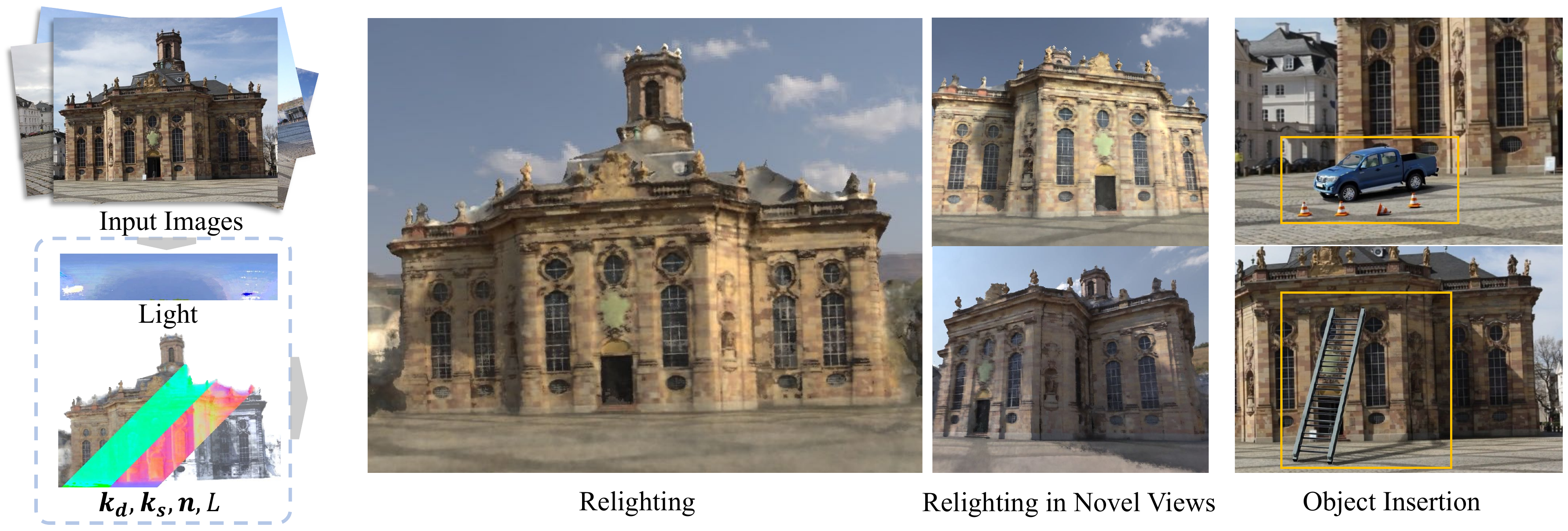}
\vspace{-2mm}
\captionof{figure}{
\small We present {\ourmodel}, an approach for reconstructing scene geometry and recovering intrinsic properties of the scene from posed camera images. Our approach works both for single and multi-illumination captured data. {\ourmodel} enables various downstream applications such as VR and AR where users may want to control the lighting of the environment and insert desired 3D objects into the scene. 
}
\end{center}
}]

\begin{abstract}
Reconstruction and intrinsic decomposition of scenes from captured imagery would enable many applications such as relighting and virtual object insertion. Recent NeRF based methods achieve impressive fidelity of 3D reconstruction, but bake the lighting and shadows into the radiance field, while mesh-based methods that facilitate intrinsic decomposition through differentiable rendering have not yet scaled to the complexity and scale of outdoor scenes. We present a novel inverse rendering framework for large urban scenes capable of jointly reconstructing the scene geometry, spatially-varying materials, and HDR lighting from a set of posed RGB images with optional depth. Specifically, we use a neural field to account for the primary rays, and use an explicit mesh (reconstructed from the underlying neural field) for modeling secondary rays that produce higher-order lighting effects such as cast shadows. By faithfully disentangling complex geometry and materials from lighting effects, our method enables photorealistic relighting with specular and shadow effects on several outdoor datasets. Moreover, it supports physics-based scene manipulations such as virtual object insertion with ray-traced shadow casting.
\end{abstract}

\vspace{-3mm}
\section{Introduction}
\label{sec:intro}

Reconstructing high fidelity 3D scenes from captured imagery is an important 
utility of scaleable 3D content creation. However, for the reconstructed environments to serve as ``digital twins'' for downstream applications such as augmented reality and gaming, we require that these environments are compatible with modern graphics pipeline and can be rendered with user-specified lighting. This means that we not only need to reconstruct 3D geometry and texture but also recover the intrinsic properties of the scene such as material properties and lighting information. This is an ill-posed, challenging problem oftentimes referred to as inverse rendering~\cite{barrow1978recovering}. 

Neural radiance fields (NeRFs)~\cite{mildenhall2020nerf} have recently emerged as a powerful neural reconstruction approach that enables photo-realistic novel-view synthesis. NeRFs can be reconstructed from a set of posed camera images in a matter of minutes~\cite{mueller2022instant,yu2022plenoxels, SunSC22dvoxgo} and have been shown to scale to room-level scenes and beyond~\cite{tancik2022blocknerf, Turki2022meganerf, xiangli2022bungeenerf}, making them an attractive representation for augmented/virtual reality and generation of digital twins. However, in NeRF, the intrinsic properties of the scene are not separated from the effect of incident light. As a result, novel views can only be synthesised under fixed lighting conditions present in the input images, i.e. a NeRF cannot be relighted~\cite{nerv2021}. 

While NeRF can be extended into a full inverse rendering formulation~\cite{bi2020neural}, this requires computing the volume rendering integral when tracing multiple ray bounces. This quickly becomes intractable due to the underlying volumetric representation. Specifically, in order to estimate the secondary rays, the volumetric density field of NeRF would have to be queried along the path from each surface point to all the light sources, scaling with $\mathcal{O}(nm)$ per point, where $n$ denotes the number of samples along each ray and $m$ is the number of light sources or Monte Carlo (MC) samples in the case of global illumination. To restrict the incurred computational cost, prior works have mostly focused on the single object setting and often assume a single (known) illumination source~\cite{nerv2021}. Additionally, they forgo the volumetric rendering of secondary rays and instead approximate the direct/indirect lighting through a visibility MLP~\cite{nerv2021, zhang2022invrender}. 

In contrast to NeRF, the explicit mesh-based representation allows for very efficient rendering. With a known mesh topology, the estimation of both primary and secondary rays is carried out using ray-mesh intersection ($\mathcal{O}(m)$) queries that can be efficiently computed using highly-optimized libraries such as OptiX~\cite{parker2010optix}. However, inverse rendering methods based on explicit mesh representations either assume a fixed mesh topology~\cite{chen2021dibrpp}, or recover the surface mesh via an SDF defined on a volumetric grid~\cite{munkberg2021nvdiffrec} and
are thus bounded by the grid resolution. Insofar, these methods have been shown to produce high-quality results only for the smaller, object-centric scenes.

In this work, we combine the advantages of the neural field (NeRF) and explicit (mesh) representations and propose {\ourmodel}\footnote{Abbreviation {\ourmodel} is derived from \textit{neural \textbf{F}ields meet \textbf{E}xplicit \textbf{G}eometric \textbf{R}epresentations} and is pronounced as \textit{"figure"}.}, a new hybrid-rendering pipeline for inverse rendering of large urban scenes. Specifically, we represent the intrinsic properties of the scene using a neural field and estimate the primary rays (G-buffer) with volumetric rendering. To model the secondary rays that produce higher-order lighting effects such as specular highlights and cast shadows, we convert the neural field to an explicit representation and preform physics-based rendering. The underlying neural field enables us to represent high-resolution details, while ray tracing secondary rays using the explicit mesh reduces the computational complexity. The proposed hybrid-rendering is fully differentiable an can be embedded into an optimization scheme that allows us to estimate 3D spatially-varying material properties, geometry, and HDR lighting of the scene from a set of posed camera images\footnote{We can also integrate depth information, if available, to further constrain the solution space.}. By modeling the HDR properties of the scene, our representation is also well suited for AR applications such as virtual object insertion that require spatially-varying lighting to cast shadows in a physically correct way.

We summarize our contributions as follows:\\[-6mm]
\begin{itemize}[leftmargin=*]
\setlength\itemsep{0.0em}
\item We propose a novel neural field representation that decomposes scene into geometry, spatially varying materials, and HDR lighting. 
\item To achieve efficient ray-tracing within a neural scene representation, we introduce a hybrid renderer that renders primary rays through volumetric rendering, and models the secondary rays using physics-based rendering. This enables high-quality inverse rendering of large urban scenes. 
\item We model the HDR lighting and material properties of the scene, making our representation well suited for downstream applications such as relighting and virtual object insertion with cast shadows.
\end{itemize}
 
{\ourmodel} significantly outperforms state-of-the-art in terms of novel-view synthesis under varying lighting conditions on the NeRF-OSR dataset~\cite{rudnev2022nerfosr}. We also show qualitative results on a single-illumination capture of an urban environment, collected by an autonomous vehicle. Moreover, we show the intrinsic rendering results, and showcase virtual object insertion as an application. Finally, we conduct a user study, in which the results of our method are significantly preferred to those of the baselines.


\section{Related Work}

\paragraph{Inverse Rendering} is a fundamental task in computer vision. The seminal work by Barrow and Tenenbaum~\cite{barrow1978recovering} aimed to understand the intrinsic scene properties including reflectance, lighting, and geometry from captured imagery. Considering the ill-posed nature~\cite{land1971lightness1} of this challenging task, early works resided to tackle the subtask known as intrinsic image decomposition, that aims to decompose an image into diffuse albedo and shading. These methods are mostly optimization-based and rely on hand-crafted priors \cite{land1971lightness1,grosse2009ground,bousseau2009user,zhao2012closed}. 
In the deep learning era, learning-based methods~\cite{bell2014intrinsic,kovacs2017shading,li2018cgintrinsics,neuralSengupta19,yu19inverserendernet,li2020inverse,li2020openrooms,Boss2020-TwoShotShapeAndBrdf,wang2021learning,wang2022neural,wimbauer2022rendering} replaced the classic optimisation pipeline and learn the intrinsic decomposition in a data-driven manner, but typically require ground truth supervision. However, acquiring ground truth intrinsic decomposition in the real world is extremely challenging. Learning-based methods thus often train on synthetic datasets~\cite{li2018cgintrinsics,neuralSengupta19,li2020inverse,li2020openrooms,wang2021learning}, and may suffer from a domain gap between synthetic and real captures. 
In addition, these methods are limited to 2.5D prediction, \ie 2D intrinsic images and a normal map, thus are unable to reconstruct the full 3D scene. 
Recent advances in differentiable rendering \cite{NimierDavidVicini2019Mitsuba2} and neural volume rendering \cite{mildenhall2020nerf} revive the optimization paradigm by enabling direct optimization of the 3D scene representation~\cite{bi2020deep,boss2021nerd,physg2021,boss2021neuralpil,zhang2021nerfactor,iron-2022,kuang2021neroic,munkberg2021nvdiffrec,hasselgren2022nvdiffrecmc,boss2022-samurai,neural_outdoor_rerender}. However, these works mostly focus on a single object setting and ignore higher-order lighting effects such as cast shadows.


\begin{figure*}[t]
\vspace{-2mm}
\centering
\includegraphics[width=1.0\linewidth,trim=0 0 0 0,clip]{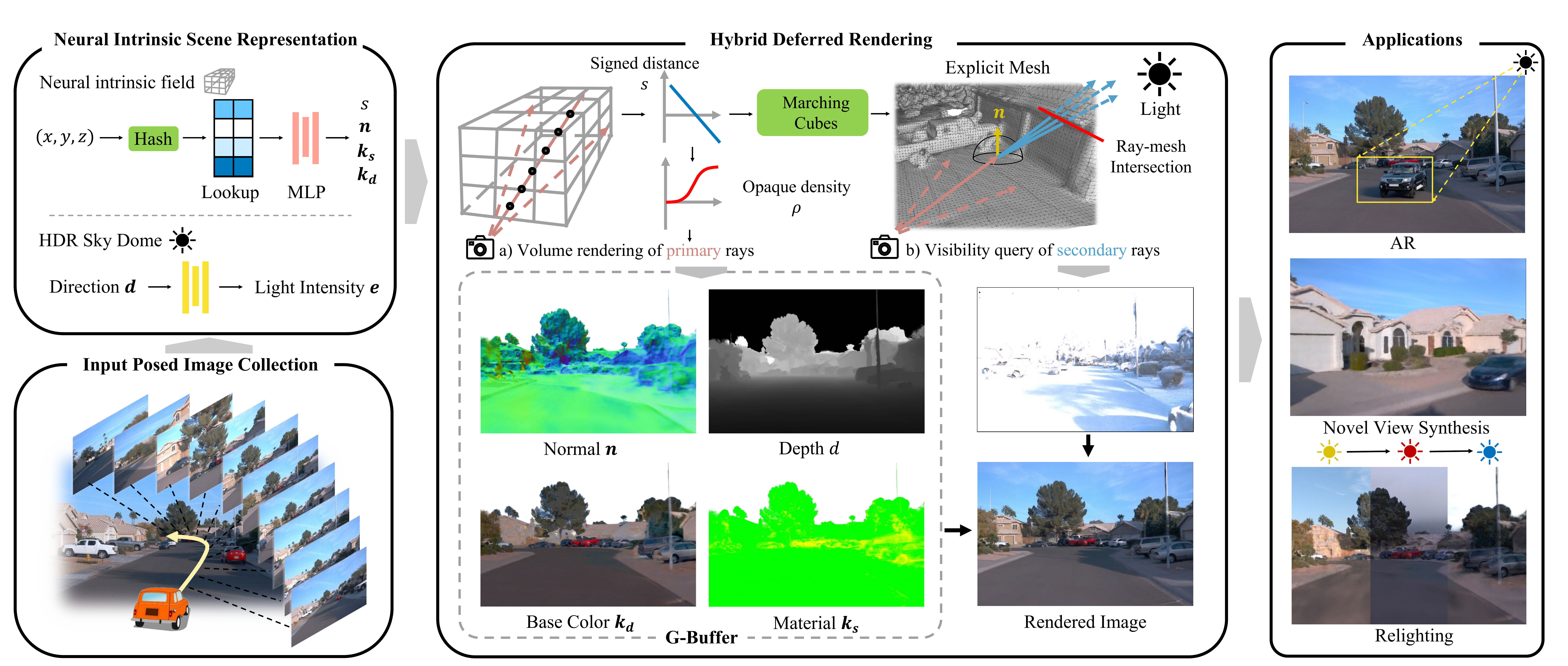}
\vspace{-6mm}
\caption{\textbf{Overview of FEGR}. Given a set of posed camera images, FEGR estimates the geometry, spatially varying materials, and HDR lighting of the underlying scene. We model the intrinsic properties of the scene using a neural intrinsic field and use an HDR Sky Dome to represent the lighting. Our Hybrid Deferred Renderer models the \textcolor{primary_pink}{\textbf{primary rays}} with volumetric rendering of the neural field, while the \textcolor{secondary_blue}{\textbf{secondary rays}} are ray-traced using an explicit mesh reconstructed from the SD field. By modeling the HDR properties of the scene FEGR can support several scene manipulations including novel-view synthesis, scene relighting, and AR.
}
\label{fig:overview}
\vspace{-3mm}
\end{figure*}

\vspace{-3mm}
\paragraph{Neural Scene Representation}
for inverse rendering mostly falls into two categories: explicit textured mesh~\cite{DIBR19,nimier2021material,chen2021dibrpp,zhang2021ners,munkberg2021nvdiffrec,hasselgren2022nvdiffrecmc} and neural fields \cite{bi2020deep,boss2021nerd,physg2021,zhang2021nerfactor,iron-2022}. 
Explicit mesh representations \cite{munkberg2021nvdiffrec,hasselgren2022nvdiffrecmc} are compatible with graphics pipeline and naturally benefit from classic graphics techniques. 
These methods show impressive performance under single-object setting but suffers from bounded resolution when scaling up to a larger scene extent. 
With the impressive image synthesis quality demonstrated by neural fields~\cite{mildenhall2020nerf}, 
recent works on inverse rendering also adopt neural fields as representation for scene intrinsic properties \cite{bi2020deep,boss2021nerd,physg2021,zhang2021nerfactor,iron-2022,verbin2022refnerf,yao2022neilf}. 
Despite the impressive results for primary ray appearance, it remains an open challenge for neural fields to represent higher order lighting effects such as cast shadows via ray-tracing. 
To reduce the complexity of ray-tracing in neural fields, prior works explore using MLP to encode visibility field \cite{nerv2021,neulighting} or Spherical Gaussian visibility~\cite{zhang2022invrender}, but typically limited to object-level or low-frequency effects. 
The closest setup to our work is NeRF-OSR \cite{rudnev2022nerfosr} that works on outdoor scene-level inverse rendering. It uses a network to represent shadows and relys on multiple illumination to disentangle shadows from albedo, but usually cannot recover sharp shadow boundaries. 
Related to our work are also methods that factorize the appearance changes through latent codes~\cite{martinbrualla2020nerfw,liu2020factorize}. These methods can modify scene appearance by interpolation of the latent codes, but do not offer explicit control of lighting conditions.

\vspace{-3mm}
\paragraph{Lighting Estimation} is a subtask of inverse rendering which aims to understand the lighting distribution across the scene, typically with the goal of photorealistic virtual object insertion. 
Existing work on lighting estimation is usually learning-based, adopting feed-forward neural networks given the input of a single image~\cite{li2018learning,legendre2019deeplight,garon2019fast,hold2017deep,hold2019deep,zhu2021cvpr,wang2022neural,SOLD-Net}. For outdoor scenes, prior work investigates network designs to predict lighting representations such as a HDR sky model~\cite{hold2017deep,zhang2019all,hold2019deep}, spatially-varying environment map~\cite{zhu2021cvpr,SOLD-Net} and a lighting volume~\cite{wang2022neural}. The key challenge for outdoor lighting estimation is to correctly estimate the peak direction and intensity of the sky, which is usually the location of the sun. This is a challenging ill-posed task where a single image input may be insufficient to produce accurate results. 
Recent optimization-based inverse rendering works jointly optimize lighting from multi-view images~\cite{boss2021nerd,physg2021,munkberg2021nvdiffrec,hasselgren2022nvdiffrecmc}, however their primary purpose of lighting is to serve the joint optimization framework for recovering material properties. Lighting representations are usually point light~\cite{nerv2021} and low frequency spherical lobes~\cite{boss2021nerd,physg2021,rudnev2022nerfosr}, which are not suited for AR applications. 
In our work, we investigate optimization-based lighting estimation to directly optimize HDR lighting from visual cues in the input imagery, such as shadows. Our neural lighting representation is used as the light source for inserting virtual objects.

\section{Method}
\label{sec:method}

Given a set of posed camera images $\{ \image_i, \camera_i  \}_{i=1}^{N_\text{RGB}}$, where $\image \in \mathbb{R}^{h \times w \times 3}$ is an image and $\camera \in \text{SE}(3)$ is its corresponding camera pose, we aim to estimate the geometry, spatially varying materials, and HDR lighting of the underlying scene. We represent the intrinsic scene properties 
using a neural field (\cref{sec:scene_representation}) and render the views with a differentiable hybrid renderer (\cref{sec:hybrid_rendering}). To estimate the parameters of the neural field, we minimize the reconstruction error on the observed views and employ several regularization terms to constrain the highly ill-posed nature of the problem  (\cref{sec:optimization}). Implementation details are provided in the Appendix. 

Note that our method addresses both single- and multi-illumination intrinsic decomposition. Existing literature~\cite{rudnev2022nerfosr} considers the multi-illumination setting that efficiently constraints the solution space of intrinsic properties and thus leads to a more faithful decomposition, while our formulation is general, and we demonstrate its effectiveness even when in the case of a single illumination capture. In the following, we keep the writing general, and address the distinction where required.  

\subsection{Neural Intrinsic Scene Representation} 
\label{sec:scene_representation}
\paragraph{Neural intrinsic field} We represent the intrinsic properties of the scene as a neural field $F_\phi: \mathbf{x} \mapsto (s,\mathbf{n}, \mathbf{k}_d, \mathbf{k}_s)$ that maps each 3D location $\mathbf{x} \in \mathbb{R}^3$ to its Signed Distance (SD) value $s \in \mathbb{R}$, normal vector  $\mathbf{n} \in \mathbb{R}^3$, base color $\mathbf{k}_d \in \mathbb{R}^3$, and materials $\mathbf{k}_s \in \mathbb{R}^2$. Here, we use $\mathbf{k}_s$ to denote the roughness and metallic parameters of the physics-based (PBR) material model from Disney~\cite{Burley2012PhysicallyBasedSA}. 
In practice we represent the neural field $F_\phi$ with three neural networks $s = f_\text{SDF}(\mathbf{x}; \bm{\theta}_\text{SDF})$, $\mathbf{n} = f_\text{norm.}(\mathbf{x}; \bm{\theta}_\text{norm.})$, and $(\mathbf{k}_d, \mathbf{k}_s) = f_\text{mat.}(\mathbf{x}; \bm{\theta}_\text{mat.})$ which are all Multi-Layer Perceptrons (MLPs) with a multi-resolution hash positional encoding~\cite{mueller2022instant}.

\vspace{-3mm}
\paragraph{HDR sky dome}
In urban scenes, the main source of light is the sky. We therefore model the lighting as an HDR environment map located at infinity, which we represent as a neural network $\mathbf{e} = f_\text{env.} (\mathbf{d}; \bm{\theta}_\text{env.})$, that maps the direction vector $\mathbf{d} \in \mathbb{R}^2$ to the HDR light intensity value $\mathbf{e} \in \mathbb{R}^3$. Specifically, $f_\text{env.}$ is again an MLP with hash positional encoding. The HDR representation of the environment map allows to perform scene manipulations such as relighting and virtual object insertion with ray-traced shadow casting.

\vspace{-3mm}
\paragraph{Single vs multi-illumination setting}
As the intrinsic properties of the scene do not change with the illumination, we use a single neural field representation of the underlying scene, and  use $M$ HDR sky maps to represent $M$ different illumination conditions present in the captured imagery.

\subsection{Hybrid Deferred Rendering}
\label{sec:hybrid_rendering}
We now describe how the estimated intrinsic properties and lighting parameters are utilized in the proposed hybrid deferred rendering pipeline. We start from the non-emissive rendering equation~\cite{kajiya1986rendering}:
\begin{equation}
	\light_o(\mathbf{x}, \lightdir_o) = \int_{\Omega}\brdf( \bm{x},  \lightdir_o, \lightdir_i)\light_i( \bm{x},  \lightdir_i) \left | \normal \cdot  \lightdir_i \right | d \lightdir_i,
	\label{eq:rendering_equation}
\end{equation}
where the outgoing radiance $\light_o$ at the surface point $\bm{x}$ and direction $ \lightdir_o$ is computed as the integral of the surface BRDF $\brdf( \bm{x},  \lightdir_o, \lightdir_i)$ multiplied by the incoming light $\light_i( \bm{x},  \lightdir_i)$  and cosine term $\left | \lightdir_i\cdot \normal \right | $, over the hemisphere $\Omega$. In all experiments 
we assume the simplified Disney BRDF model.

Albeit an accurate model, the rendering equation does not admit an analytical solution and is therefore commonly solved using MC methods. However, due to the volumetric nature of our scene representation, sampling enough rays to estimate the integral quickly becomes intractable, even when relying on importance sampling. To alleviate the cost of evaluating the rendering equation, while keeping the high-resolution of the volumetric neural field, we propose a novel hybrid deferred rendering pipeline. 
Specifically, we first use the neural field to perform volumetric rendering of primary rays into a G-buffer that includes the surface normal, base color, and material parameters for each pixel. We then extract the mesh from the underlying SD field, and perform the shading pass in which we compute illumination by integrating over the hemisphere at the shading point using MC ray tracing. This allows us to synthesize high quality shading effects, including specular highlights and shadows.

\vspace{-3mm}
\paragraph{Neural G-buffer rendering}
To perform volume rendering of the G-buffer $\mathbf{G} \in \mathbb{R}^{h \times w \times 8}$, which contains a normal map $\normalmap \in \mathbb{R}^{h \times w \times 3}$, a base color map $\albedomap \in \mathbb{R}^{h \times w \times 3}$, a material map $\materialmap \in \mathbb{R}^{h \times w \times 2}$ and a depth map $\depthmap \in \mathbb{R}^{h \times w}$, we follow the standard NeRF volumetric rendering equation~\cite{zhang2020nerf}. For example, consider base color $\albedo$ and let $\ray = \origin  + t\dir$ denote the camera ray with origin $\origin$ and direction $\dir$. The alpha-composited base color map $\albedomap$ along the ray can then be estimated as 
\begin{equation}
     \albedomap(\ray)  = \int_{t_n}^{t_f} T(t) \density(\ray(t)) \albedo(\ray(t)) dt,
     \label{eq:vol_rendering}
\end{equation}
where $T(t) = \exp{(-\int_{t_n}^{t}\density(\ray(s))ds)}$ denotes the accumulated transmittance, and $t_n$, $t_f$ are the near and far bound respectively. Following~\cite{wang2021neus} the opaque density $\density(t)$ can be recovered from the underlying SD field as:

\begin{equation}
\density(\ray(t)) = \max(\frac{-\frac{\text{d}\Phi_{\kappa}}{\text{d}t}(f_\text{SDF}(\ray(t)))}{\Phi_{\kappa}(f_\text{SDF}(\ray(t)))},0) 
\end{equation}
where $\Phi_{\kappa}(x) = \text{Sigmoid}(\kappa x)$ and $\kappa$ is a learnable parameter~\cite{wang2021neus}. The surface normals and material buffer of $\mathbf{G}$ are rendered analogously.  We render the depth buffer $\depthmap \in \mathbb{R}^{h \times w}$ as radial distance:
\begin{equation}
     \depthmap(\ray)  = \int_{t_n}^{t_f} T(t) \density(\ray(t)) t dt,
\end{equation}

\vspace{-3mm}
\paragraph{Shading pass}
Given the G-buffer, we can now perform the shading pass. To this end, we first extract an explicit mesh $\mesh$ of the scene from the optimized SD field using marching cubes~\cite{Lorensen87marchingcubes}. We then estimate Eq.~\eqref{eq:rendering_equation} based on the rendered G-buffer, Specifically, for each pixel in the G-buffer, we query its intrinsic parameters (surface normal, base color, and material) and use the depth value to compute its corresponding 3D surface point $\mathbf{x}$. We then perform MC sampling of the secondary rays from the surface point $\mathbf{x}$. While previous work assume a simplified case where all the rays reach the light source~\cite{chen2021dibrpp,physg2021,munkberg2021nvdiffrec}, the extracted mesh $\mesh$ enables us to determine the visibility $\visibility$  of each secondary ray with OptiX~\cite{parker2010optix}, a highly-optimized library for ray-mesh intersection queries. Here, the $\visibility$ is defined as:
\begin{align}
	\visibility_i(x, \lightdir_i, \mesh) = \begin{cases} 0 & \text{if $ \lightdir_i$ is blocked by $\mesh$} \\
		1 & \text{otherwise} \end{cases}
\end{align}

The visibility of each ray is incorporated into the estimation of the incoming light as $\light_i(x, \lightdir_i) = \visibility_i(x, \lightdir_i, \mesh) f_\text{env.}(\lightdir_i; \bm{\theta}_\text{env.})$. Explicit modeling of the visibility in combination with the physically based BRDF enables us to compute higher-order lighting effects such as cast shadows. 

In practice, we trace 512 secondary rays by importance sampling the BSDF and the HDR environment map. Following~ \cite{hasselgren2022nvdiffrecmc}, we combine samples of the two sampling strategies using multiple importance sampling \cite{veach1995optimally}. Using the highly optimized library OptiX, ray-tracing of the secondary rays is carried out in real-time. Once our representation is optimized, we can export the environment map $\mathbf{E} \in \mathbb{R}^{h_\text{e} \times w_\text{e} \times 3}$ (evaluating $f_\text{env.}$ once per each texel of $\mathbf{E}$), allowing us to perform importance sampling using $\mathbf{E}$ without additional evaluations of $f_\text{env.}$. 
During optimization, when the SD field is continuously updated, we reconstruct a new explicit mesh every 20 iterations. Empirically, this offers a good compromise between the rendering quality and efficiency.

\subsection{Optimizing the Neural Scene Representation}
\label{sec:optimization}
Given a set of posed images captured under unknown illumination condition and, when available, LiDAR point clouds, we optimize the neural scene representation end-to-end by minimizing the loss:
\begin{align}
    \mathcal{L} =& \mathcal{L}_{\text{render}} + \lambda_\text{depth}\mathcal{L}_{\text{depth}} + \lambda_\text{rad.}\mathcal{L}_{\text{rad.}} + \lambda_\text{norm.}\mathcal{L}_{\text{norm.}} \nonumber \\
    &+ \lambda_\text{shade}\mathcal{L}_{\text{shade}} + \lambda_\text{reg.}\mathcal{L}_{\text{reg.}}, 
\end{align}
where $\mathcal{L}_{\text{render}}$, $\mathcal{L}_{\text{depth}}$ are the reconstruction loss on the observed pixel and LiDAR rays and $\mathcal{L}_{\text{rad.}}$, $\mathcal{L}_{\text{norm.}}$, and $\mathcal{L}_{\text{shade}}$ are used to regularize the geometry, normal field, and lighting, respectively. We additionally employ several regularization terms $\mathcal{L}_{\text{reg.}}$ to constrain the ill-posed nature of the problem. $\lambda_*$ are the weights used to balance the contribution of the individual terms. More details are discussed in the Appendix. 

\vspace{-3mm}
\paragraph{Rendering loss}
As the main supervision signal, we use the L1 reconstruction loss between input images and corresponding views rendered using the proposed hybrid renderer:
\begin{equation}
    \mathcal{L}_{\text{render}} = \frac{1}{|\mathcal{R}|} \sum_{\mathbf{r} \in \mathcal{R}} |C_{\text{render}}(\mathbf{r}) - {C}_\text{gt}(\mathbf{r})|, 
\end{equation}
where $C_{\text{render}}(\mathbf{r})$ denotes the rendered RGB value for the camera ray $\mathbf{r}$, ${C}_\text{gt}(\mathbf{r})$ is the ground truth RGB value of the corresponding ray, and $\mathcal{R}$ denotes the set of camera rays in a single batch. As our representation is fully differentiable, the gradients of  $\mathcal{L}_{\text{render}}$ are propagated to all intrinsic properties in the neural field, as well as to the HDR sky map. 

\vspace{-3mm}
\paragraph{Geometry supervision} 
To regularize the underlying SD field to learn reasonable geometry, we introduce an auxiliary radiance field $C_\text{rad.} = f_{\text{rad.}}(\mathbf{x}, \mathbf{d}; \bm{\theta}_{\text{rad.}})$ that maps each 3D location $\mathbf{x}$ along direction $\mathbf{d}$ to its emitted color and define the loss as
\begin{equation}
    \mathcal{L}_{\text{rad.}} = \frac{1}{|\mathcal{R}|} \sum_{\mathbf{r} \in \mathcal{R}} |C_\text{rad.}(\mathbf{r}) - {C}_\text{gt}(\mathbf{r})|,
\end{equation}
where $C_\text{rad.}(\mathbf{r})$ is the RGB color obtained through volumetric rendering\footnote{$f_{\text{rad.}}$ only encodes the radiance. The SD field used to perform volumetric rendering is shared with our neural scene representation.} 
of $f_{\text{rad.}}$ along the ray $\mathbf{r}$, $\hat{C}(\mathbf{r})$ is the corresponding ground truth RGB, and $\mathcal{R}$ denotes the set of camera rays in a single batch. Note that the radiance field $f_{\text{rad.}}$ is only used to provide an auxiliary supervision of the geometry and is discarded after the optimization converges.

For driving data where additional LiDAR measurements are available, we use L1 loss on the range value 
\begin{equation}
    \mathcal{L}_{\text{depth}} = \frac{1}{|\mathcal{R_\text{d}}|} \sum_{\mathbf{r} \in \mathcal{R_\text{d}}} |\depthmap(\ray) - {\depthmap}_\text{gt}(\ray)|.
\end{equation}

\vspace{-3mm}
\paragraph{Normal regularization} 
While the normal vector $\tilde{\mathbf{n}}_\mathbf{x}$ at the point $\mathbf{x}$ could be directly estimated from the SD field as $\tilde{\mathbf{n}}_\mathbf{x} = -\frac{\nabla_{\mathbf{x}}f_\text{SDF}}{||\nabla_{\mathbf{x}} f_\text{SDF}||}$, we empirically observe that such formulation results in smooth normal vectors that cannot represent high-frequency geometry details. Instead, we estimate the normal vectors $\mathbf{n}_\mathbf{x}$ through volumetric rendering (see \cref{sec:hybrid_rendering}) and use $\tilde{\mathbf{n}}_\mathbf{x}$ only as a regularizer in form of an angular loss
\begin{equation}
    L_{\text{norm.}} = \frac{1}{|\mathcal{R}|} \sum_{\mathbf{r} \in \mathcal{R}} \cos^{-1}{(\left | \tilde{\mathbf{n}}_\mathbf{x} \cdot \mathbf{n}_\mathbf{x} \right |)},
\end{equation}
where $\left |  \cdot \right |$ denotes the dot product. Normal vectors obtained through volumetric rendering are capable of capturing high-frequency details while also respecting the low frequency. 

\vspace{-3mm}
\paragraph{Shading regularization}

Inverse rendering under unknown illumination is a highly ill-posed problem. Without adequate regularization, optimization-based methods tend to bake shadows into diffuse albedo, rather than explaining them as a combination of geometry and environment map~\cite{hasselgren2022nvdiffrecmc}. In urban scenes, shadows are often cast on areas with a single dominant albedo, resulting in shadow boundaries that can be used as visual cues for intrinsic decomposition. In addition, these regions are often in the same semantic class, e.g., road, sidewalks, and buildings. 
Based on this observation, we introduce a set of auxiliary learnable parameters -- one albedo per semantic class, and encourage its re-rendering to be consistent with the groundtruth image: 
\begin{equation}
     \mathcal{L}_{\text{shade}} = \frac{1}{B} \sum_{b=1}^{B} \frac{1}{|\mathcal{R}_{b}|} \sum_{\mathbf{r} \in \mathcal{R}_{b}} |C_\text{diffuse}^b(\mathbf{r}) - \hat{C}(\mathbf{r})|,
\end{equation}
where $B$ is the number of semantic classes, $\mathbf{k}_{\text{sem}}^b \in \mathbb{R}^3$ is the $b$-th learnable semantic albedo, $\mathcal{R}^b$ is the set of camera rays that belong to the $b$-th semantic class, $C_\text{diffuse}^b(\mathbf{r}) = \mathbf{k}_{\text{sem}}^b \mathbf{s}_\text{diffuse}$ is the rendered color, and $\mathbf{s}_\text{diffuse}$ is the diffuse shading in deferred rendering. Intuitively, the shading regularization term encourages the optimization to explain the cast shadows by adapting the environment map, due to the limited capacity of per-semantic class albedo. The semantic segmentation are computed with an off-the-shelf semantic segmentation network~\cite{tao2020hierarchical}.

\vspace{-3mm}
\paragraph{Optimization scheme} 
Since our hybrid renderer relies on the explicit mesh extracted from the SD field, similar to NeRFactor~\cite{zhang2021nerfactor}, we first initialise the geometry by optimizing with only radiance, then optimize with other scene intrinsics using all losses. More details are in the Appendix.


\section{Experiments}
\label{sec:results}
\begin{table}[t]
    \setlength{\tabcolsep}{4pt}
    \renewcommand{\arraystretch}{1.2}
	\centering
	\resizebox{\columnwidth}{!}{
    \begin{tabular}{lcc|cc|cc}
    \toprule
    & \multicolumn{2}{c|}{Site 1} & \multicolumn{2}{c|}{Site 2} & \multicolumn{2}{c}{Site 3}  \\
     & PSNR $\uparrow$ & MSE $\downarrow$ & PSNR $\uparrow$ & MSE $\downarrow$ & PSNR $\uparrow$ & MSE $\downarrow$ \\
    \midrule
    NeRF-OSR~\cite{rudnev2022nerfosr} & 19.34 & 0.012 & 16.35 & 0.027 & 15.66 & 0.029 \\
    Ours  &  \textbf{21.53} & \textbf{0.007} & \textbf{17.00} & \textbf{0.023} & \textbf{17.57} & \textbf{0.018} \\
    Ours (mesh only) & 18.94 & 0.013 & 16.50 & 0.025 & \underline{16.86} & \underline{0.021} \\
    Ours (w/o shadow) & 20.62 & 0.009 & 16.17 & 0.028 & 16.15 & 0.024 \\
    Ours (w/o exposure) & \underline{20.70} & \underline{0.009} & \underline{16.70} & \underline{0.025} & 16.09 & 0.025 \\
    \bottomrule
    \end{tabular}
    }
	\vspace{-2mm}
	\caption{Outdoor scene relighting results on \textit{NeRF-OSR} dataset.} 
	\label{tab:nerf_osr}
    \label{tab:ablation}
\end{table}


\begin{figure*}[ht!]
    \centering
    \includegraphics[width=0.95\linewidth]{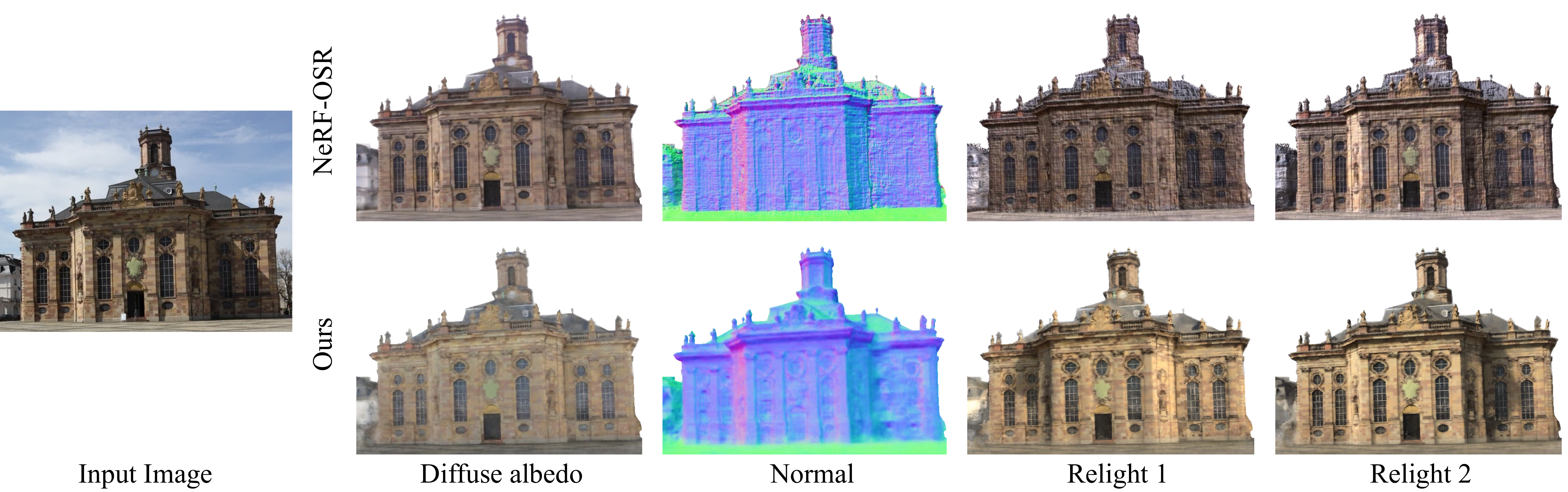}
    \vspace{-2mm}
    \caption{Qualitative results of scene relighting on \textit{NeRF-OSR}~\cite{rudnev2022nerfosr} dataset. Our method reconstructs clean diffuse albedo and enables high-quality relighting with photo-realistic cast shadow.} 
    \label{fig:qua_osr}
    \vspace{-2mm}
\end{figure*}


\begin{figure*}[t]
\centering
\includegraphics[width=0.95\linewidth]{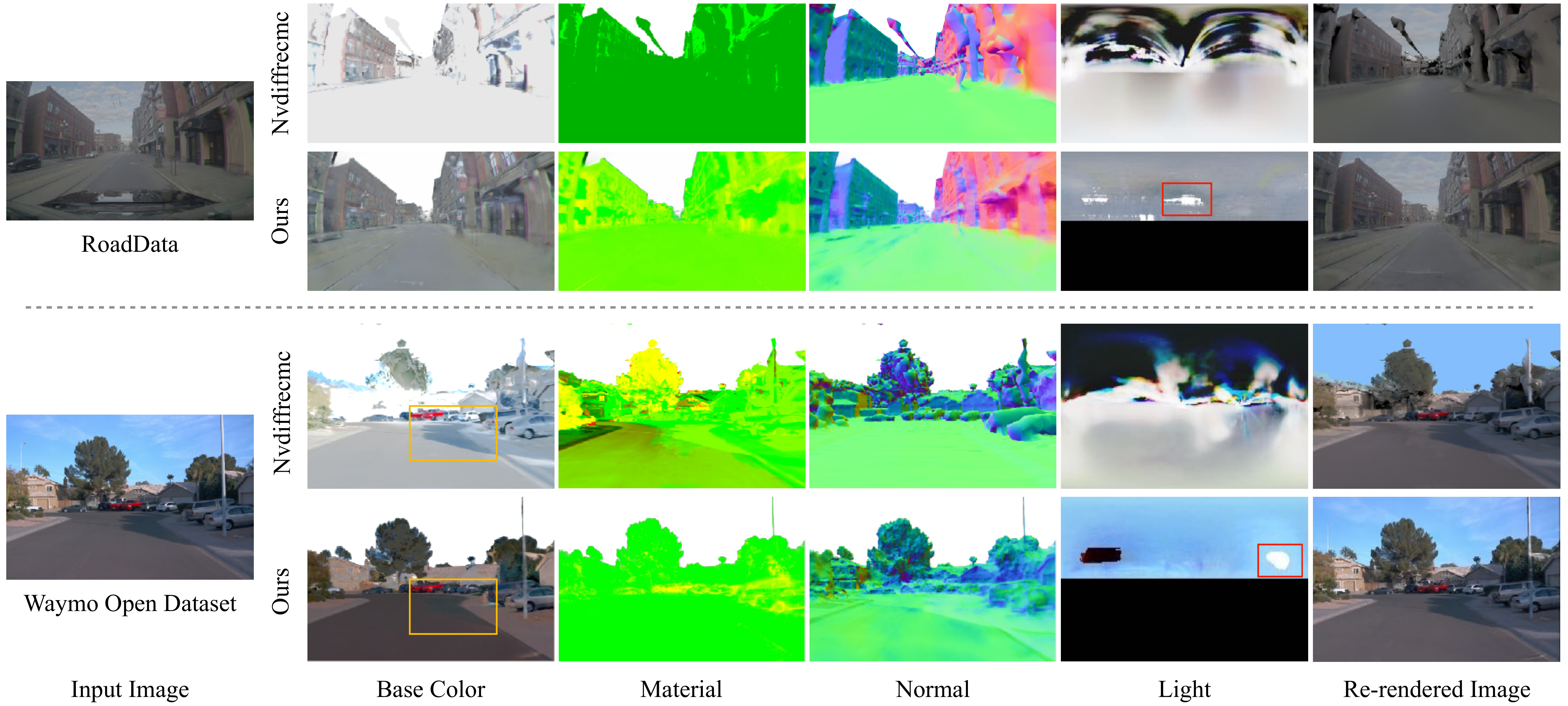}
\vspace{-3mm}
\caption{Qualitative results of intrinsic scene decomposition on the \textit{Driving} dataset. Our method successfully separates shadows from diffuse albedo (see \textcolor{mark}{mark}), and reconstructs a high intensity, small area in the environment map (see \textcolor{red}{mark}). 
}
\label{fig:qua_wod_driving}
\vspace{-2mm}
\end{figure*}

\begin{figure*}[t]
    \centering
    \includegraphics[width=0.95\linewidth]{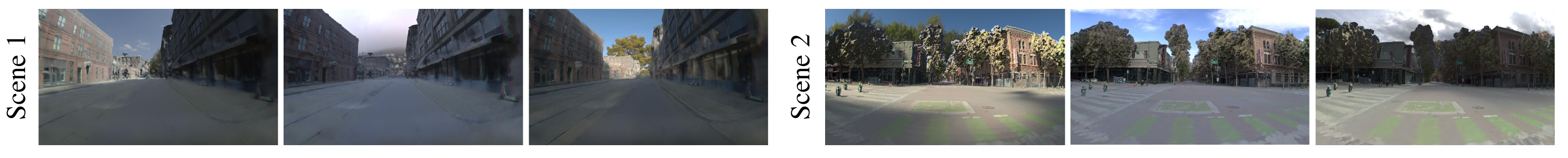}
    \vspace{-3mm}
    \caption{Qualitative results of scene relighting on \textit{RoadData}. We show 3 relighting results for each scene. } 
    \label{fig:relight_nv}
    \vspace{-2mm}
\end{figure*}

\begin{figure*}[htbp]
\centering
\begingroup
\setlength{\tabcolsep}{0.5pt}
\resizebox{0.95\linewidth}{!}{
\begin{tabular}{cccc}
\includegraphics[width=0.3\linewidth]{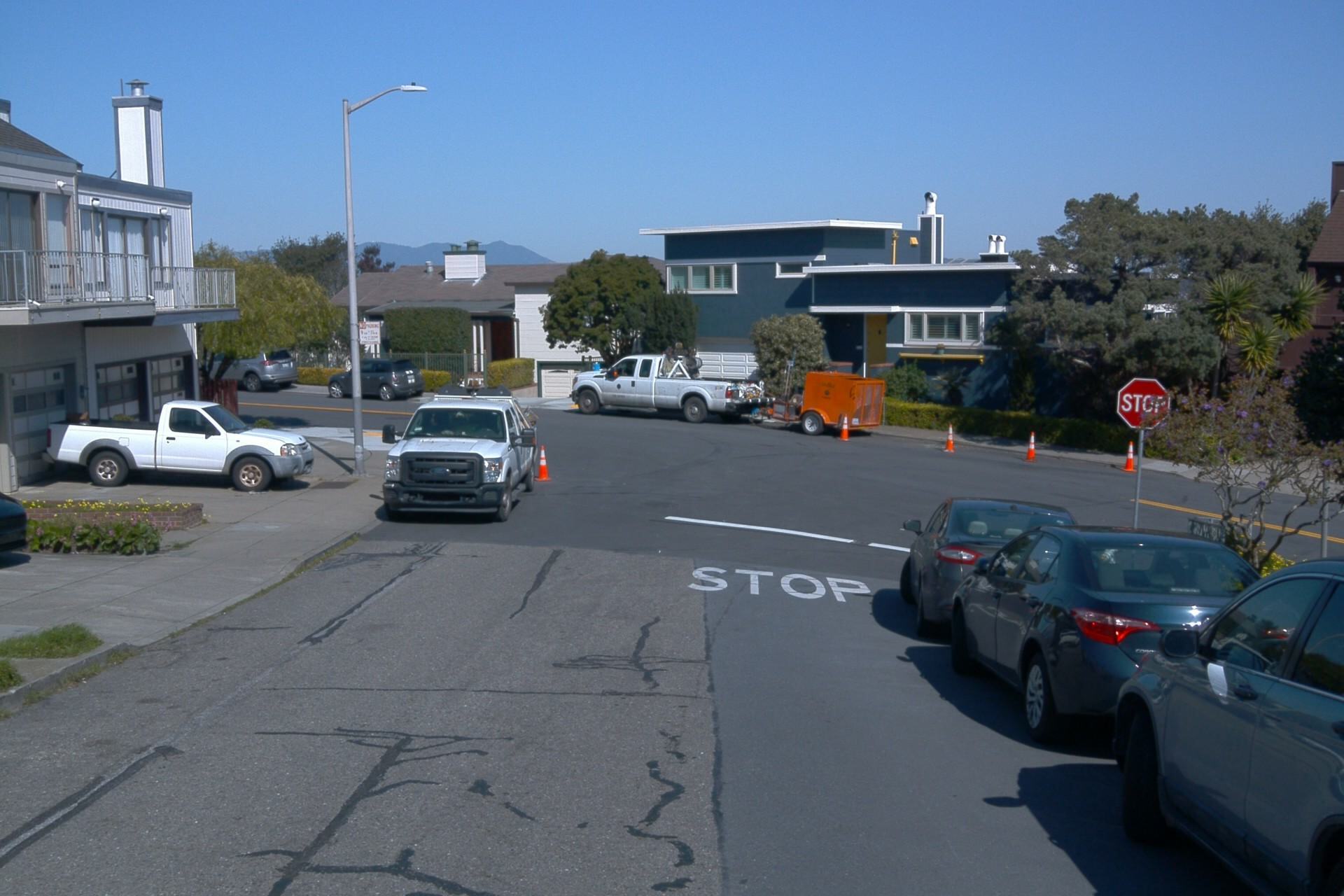} & 
\includegraphics[width=0.3\linewidth]{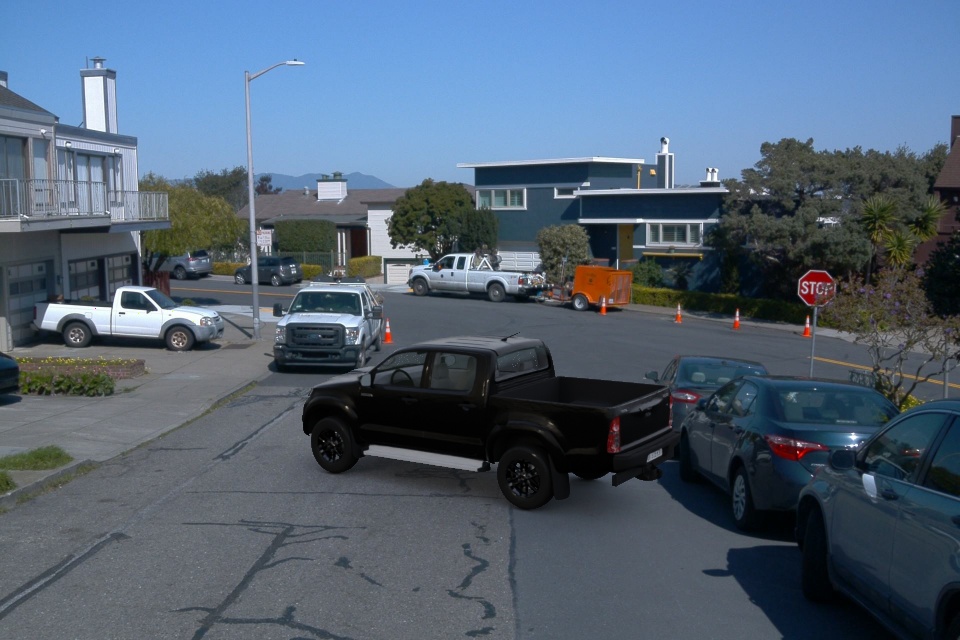} & 
\includegraphics[width=0.3\linewidth]{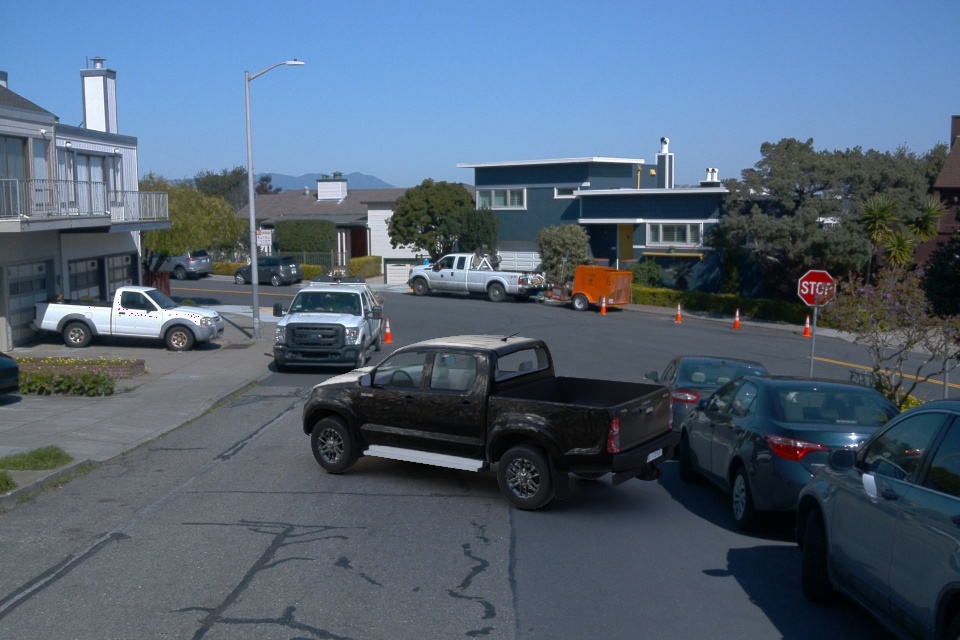} & 
\includegraphics[width=0.3\linewidth]{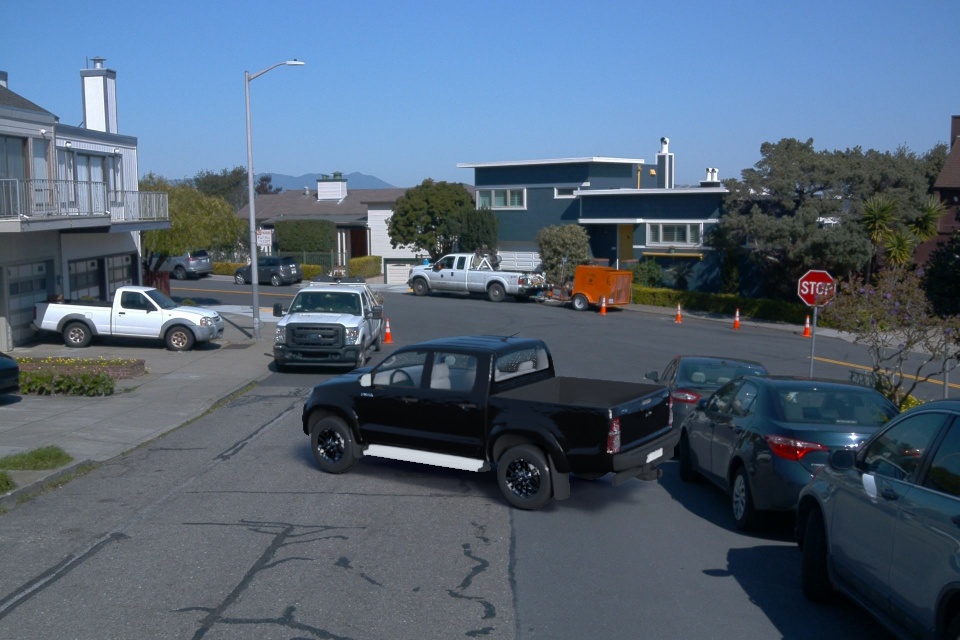} \\
\includegraphics[width=0.3\linewidth]{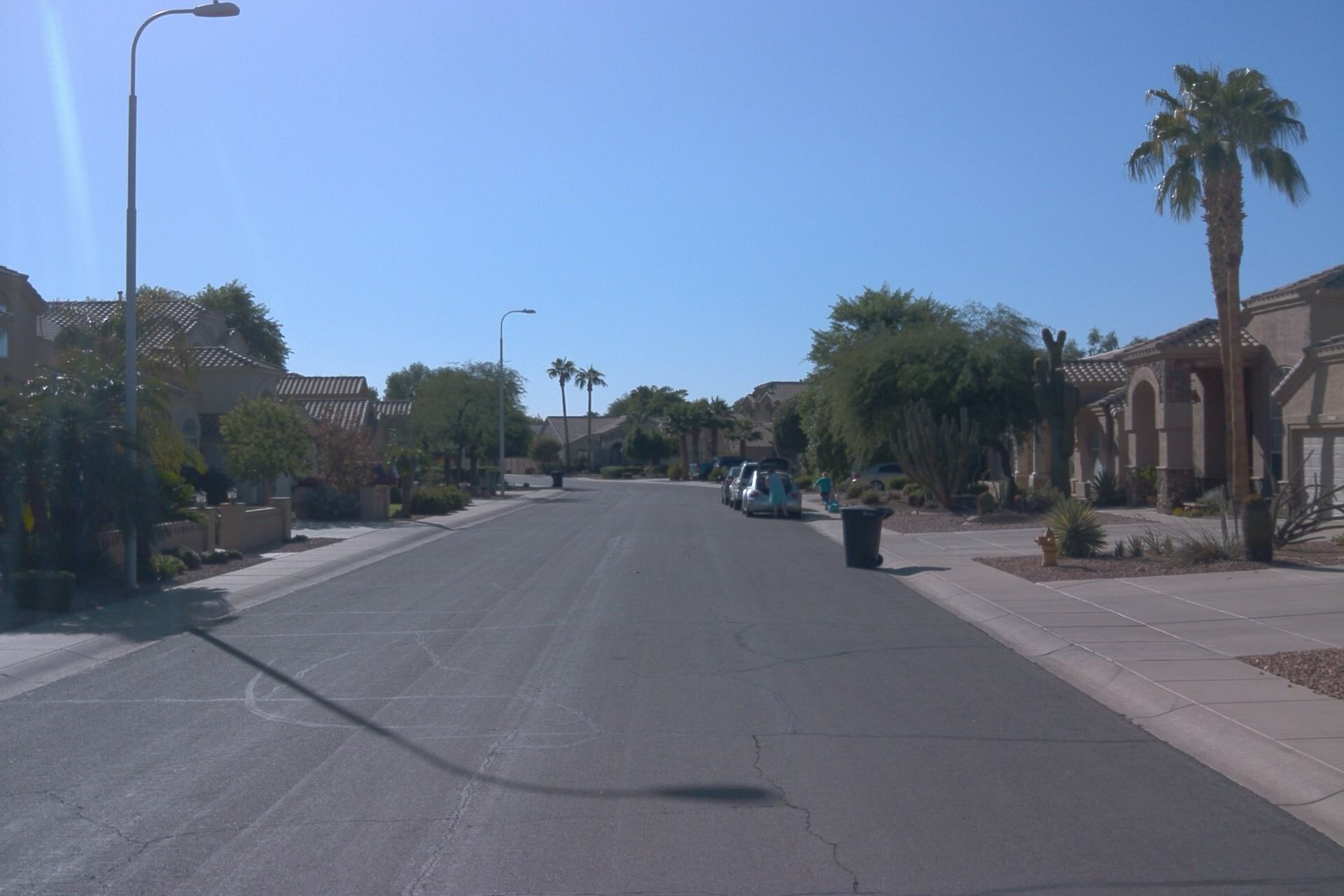} & 
\includegraphics[width=0.3\linewidth]{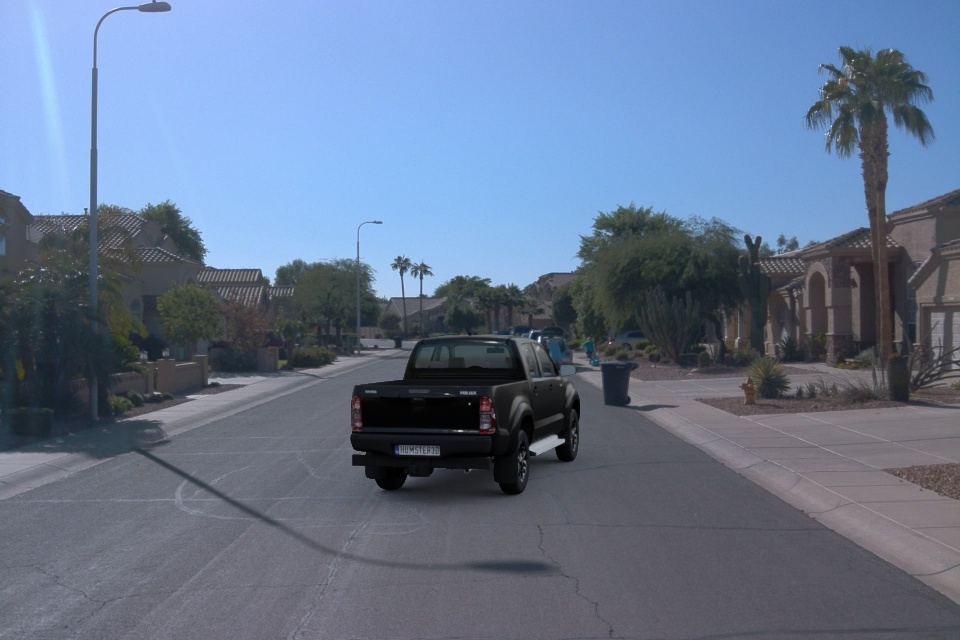} & 
\includegraphics[width=0.3\linewidth]{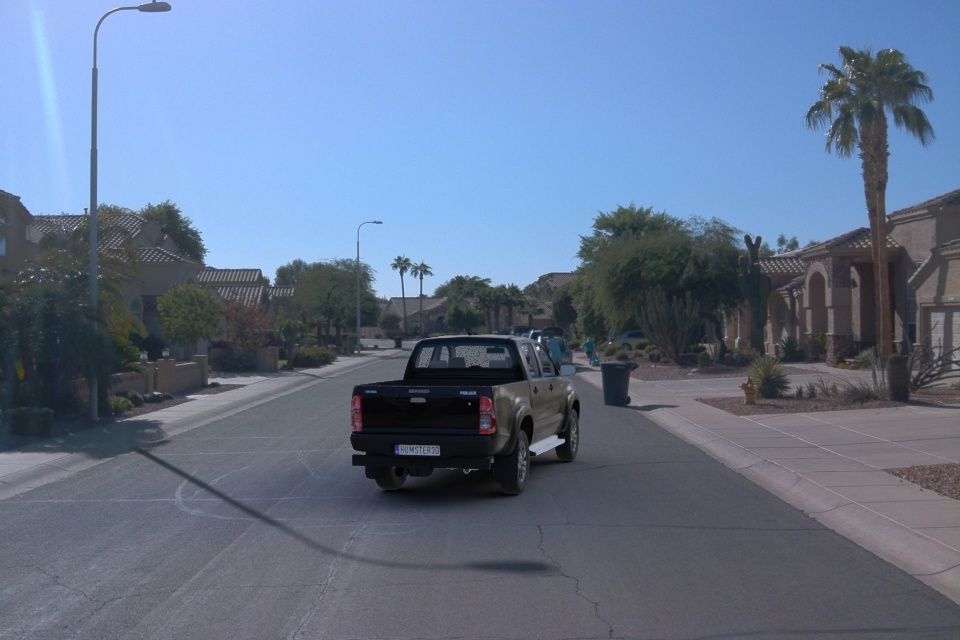} & 
\includegraphics[width=0.3\linewidth]{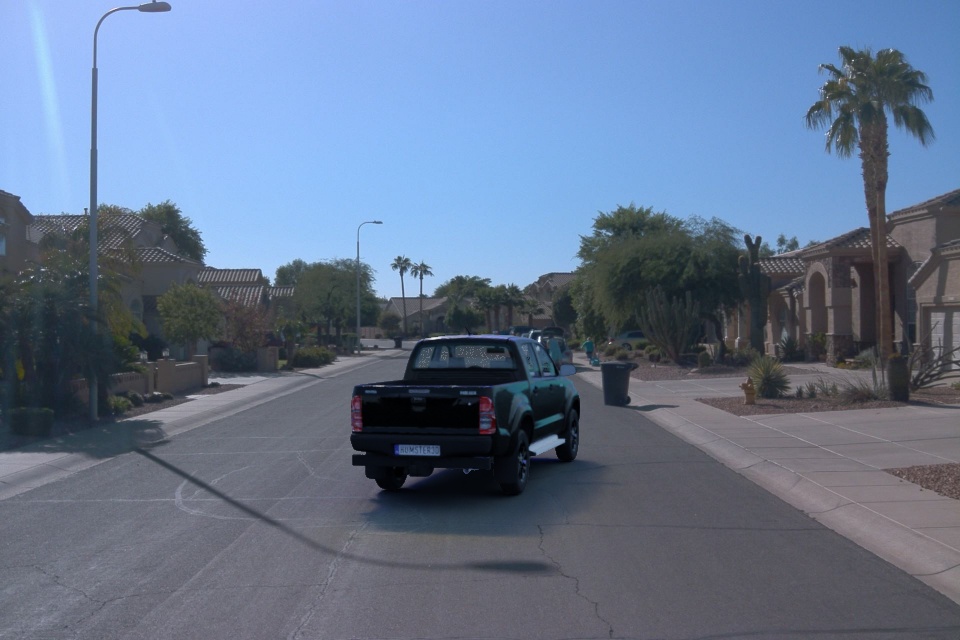} \\
Input image & Hold-Geoffroy \etal~\cite{hold2019deep} & Wang \etal~\cite{wang2022neural} & Ours \\
\end{tabular}
}
\endgroup
\vspace{-2mm}
\caption{Qualitative comparison of virtual object insertion. Our method faithfully reconstructs the environment map and produces photo-realistic cast shadows with sharp boundaries.}
\label{fig:object_insertion}
\vspace{-3mm} 
\end{figure*}

We use three urban outdoor datasets to evaluate \ourmodel~and to justify our design choices. We start by describing the datasets and the evaluation setting used in our experiments (\cref{sec:dataset}). We then provide a quantitative and qualitative evaluation of inverse rendering of large urban scenes under multi-illumination setting (\cref{sec:benchmark}). Additionally, we evaluate our method on a very challenging scenario of autonomous driving scenes captured under a single illumination. Finally, we showcase that {\ourmodel} can support downstream tasks such as virtual object insertion with ray-traced shadow casting (\cref{sec:application}).

\subsection{Datasets and evaluation setting}
\label{sec:dataset}

\paragraph{NeRF-OSR dataset~\cite{rudnev2022nerfosr}} contains in total eight outdoor scenes captured using a DSLR camera in 110 recording sessions across all scenes. Each session also contains an environment map estimated from the images acquired using a 360$\degree$ camera. In our evaluation, we follow the setting proposed in~\cite{rudnev2022nerfosr}. Specifically, we use three scenes for quantitative evaluation and use 13/12/11 sessions respectively to optimize the parameters of our neural scene representation. We then use environment maps from five other recording sessions to relight each scene and measure average PSNR and MSE between the rendered and ground-truth images. To remove dynamic objects, sky and vegetation pixels  we again follow~\cite{rudnev2022nerfosr} and use the segmentation masks predicted by an off-the-shelf semantic segmentation network~\cite{tao2020hierarchical}.

\vspace{-3mm}
\paragraph{Driving dataset} includes two scenes captured by autonomous vehicles (AV) in an urban environment. The first scene is from the Waymo Open Dataset (WOD)~\cite{sun2020scalability}, and has a 20-second clip acquired by five pinhole cameras and one 64-beam LiDAR sensor at 10 Hz. We use all five camera views for our experiments. The second set of scenes is also from a high-quality AV dataset (dubbed RoadData) acquired in-house. It is captured using eight high resolution (3848x2168 pixel) cameras with calibrated distortion and one 128-beam LiDAR. We only use images from the front-facing 120 FoV camera. For both scenes, we additionally rely on LiDAR point clouds for depth supervision. The Driving dataset is challenging as it records large street environments with complex geometry, lighting, and occlusion, and typically with a fast camera motion. It also only records a scene in a single drive thus providing only single illumination capture. However, urban environments are of high interest to digitize so as to serve as content to a variety of downstream applications such as gaming and AV simulation.

\vspace{-3mm}
\paragraph{Baselines}
We select different baselines for each of the tasks. In the relighting benchmark on NeRF-OSR dataset, we compare our method to NeRF-OSR~\cite{rudnev2022nerfosr}. For the challenging inverse rendering problem on the Driving dataset, we compare to Nvdiffrecmc~\cite{hasselgren2022nvdiffrecmc}. Finally, we perform a user-study and compare {\ourmodel} to Hold-Geoffroy et al.~\cite{hold2019deep} and Wang et al.~\cite{wang2022neural} on the task of virtual object insertion.

\subsection{Evaluation of Inverse Rendering}
\label{sec:benchmark}

\paragraph{Outdoor scene relighting}

\cref{tab:nerf_osr} shows the quantitative evaluation of the relighting performance on the NeRF-OSR dataset. {\ourmodel} significantly outperforms the baseline across all three scenes in terms of both PSNR and MSE. In \cref{fig:qua_osr} we additionally show qualitative results obtained by relighting the scenes using two different environment maps. 
The normal vectors estimated by NeRF-OSR contain high-frequency noises, which result in artifacts when relighting the scene with strong directional light. On the other hand, {\ourmodel} succeeds in faithfully decomposing the geometry and material from lighting and yields visually much more pleasing results with sharp cast shadows and high-quality details.

\vspace{-3mm}
\paragraph{Ablation study}
We ablate our design choices by comparing {\ourmodel} to three simplified versions: (i) \textit{Ours (mesh only)} denotes a version where we transfer the intrinsic properties from the neural field to the vertices of the reconstructed mesh and compute both primary and secondary rays from the mesh representation, (ii) in \textit{Ours (w/o shadows)} we disregard secondary rays and render only the primary rays from the neural field, and (iii) \textit{Ours (w/o exposure)} where we do not perform per color channel exposure compensation (see Appendix for more details). \cref{tab:ablation} shows that the proposed combination of the high resolution neural field with the explicit mesh is crucial to high-quality results. Physically based ray-tracing of shadows and exposure compensation further boost our performance and result in gains of up to 1.5 dB PSNR.

\vspace{-3mm}
\paragraph{Driving dataset}
\textit{Driving} dataset is challenging in several aspects: 
(i) The scenes are large (up to $200 \text{m} \times 200 \text{m}$ in horizontal plane) with complex geometry and spatially-varying material; 
(ii) Environment illumination is unknown and could contain high intensity from the sun; 
(iii) Images are captured by a fast-moving vehicle ($\approx$ 10 m/s on \textit{WOD} dataset) resulting in motion blur and HDR artifacts. 
Even so, our method still achieves superior intrinsic scene decomposition on both scenes, leading to photo-realistic view-synthesis results (see \cref{fig:qua_wod_driving}). Compared to Nvdiffrecmc\footnote{In driving scenario where inputs is a restricted set of views in an "inside-looking-out" manner, Nvdiffrecmc relies heavily on depth supervision to recover the geometry. This leads to artifacts on surfaces that are not observed by LiDAR or have incorrect depth signal (e.g. windows). The erroneous geometry hurts the estimation of intrinsic properties.}~\cite{hasselgren2022nvdiffrecmc}, we reconstruct cleaner base color, more accurate geometry, and higher resolution environment maps. It is worth noting that Nvdiffrecmc is designed for "outside-looking-in" setups with 360-view coverage and does not directly work on \textit{Driving} dataset without modifications. In \cref{fig:relight_nv} we additionally show the qualitative scene relighting results under the challenging illumination settings.

\subsection{Application to virtual object insertion} 
\label{sec:application}
\paragraph{Qualitative comparison}

\cref{fig:object_insertion} shows qualitative results of object insertion on Driving dataset. {\ourmodel} is capable of faithfully representing the location of the sun, resulting in cast shadows that agree with the surroundings and yield a photo-realistic insertion.  

\vspace{-3mm}
\paragraph{User study}

\begin{table}[t!]
\centering
\resizebox{0.8\linewidth}{!}{
\addtolength{\tabcolsep}{6pt}
\begin{tabular}{l|c}
\toprule
 & \%  Ours is preferred \\
\midrule
vs Hold-Geoffroy \etal~\cite{hold2019deep} & $86.2$   \%  \\ 
vs Wang \etal~\cite{wang2022neural}      	& $68.9$   \%  \\
\bottomrule
\end{tabular}
} 
\vspace{-2mm} 
\caption{User study results of object insertion quality. Users consistently prefer ours over results from baseline methods.} 
\label{table:userstudy} 
\vspace{-3mm} 
\end{table}

To quantitatively evaluate the object insertion results of {\ourmodel} against other baselines, we conduct a user study using Amazon Mechanical Turk. In particular, we show the participants two augmented images with the same car inserted by our method and by the baseline in random order, and ask them to evaluate which one is more photo-realistic based on: (i) the quality of cast shadows, and (ii) the quality of reflections. For each baseline comparison, we invite 9 users to judge 29 examples and use the majority vote for the preference for each example. The results of the user study are presented in \cref{table:userstudy}. A significant majority of the participants agree that {\ourmodel} yields more realistic results than all baselines, indicating a more accurate lighting estimation of our method.


\section{Conclusion}
\label{sec:conc}

We introduced {\ourmodel}, a novel hybrid rendering pipeline for inverse rendering of large urban scenes. {\ourmodel} combines high-resolution of the neural fields with the efficiency of explicit mesh representations and is capable of extracting the scene geometry, spatially varying materials, and HDR lighting from a set of posed camera images. The formulation of {\ourmodel} is flexible and it supports both single and multi-illumination data. We demonstrated that {\ourmodel} consistently outperforms SoTA methods across various challenging datasets. Finally, we have demonstrated that {\ourmodel} can seamlessly support various scene manipulations including relighting and virtual object insertion (AR). 

\vspace{-4mm}
\paragraph{Limitations}
While {\ourmodel} makes an important step forward in neural rendering of large urban scenes, it naturally also has limitations. Inverse rendering is a highly-ill posed problem in which the solution spaces has to be constrained, especially when operating on single illumination data. We currently rely on manually designed priors to define regularization terms. In the future we would like to explore ways of learning these priors from the abundance of available data. Similar to most methods based on neural fields, {\ourmodel} is currently limited to static scenes. A promising extension in the future could incorporate advances in dynamic NeRFs~\cite{gao2022dynamic} to mitigate this problem.

{\small
\bibliographystyle{ieee_fullname}
\bibliography{egbib}
}

\end{document}


\title{Supplementary Material: Neural Fields meet Explicit Geometric Representations for Inverse Rendering of Urban Scenes}

\author{
  Zian Wang$^{1,2,3}$ 
  \quad Tianchang Shen$^{1,2,3}$ 
  \quad Jun Gao$^{1,2,3}$ 
  \quad Shengyu Huang$^{1,4}$ 
  \quad Jacob Munkberg$^{1}$ \\
  \quad Jon Hasselgren$^{1}$ 
  \quad Zan Gojcic$^{1}$ 
  \quad Wenzheng Chen$^{1,2,3}$ 
  \quad Sanja Fidler$^{1,2,3}$ \\
  $^1$NVIDIA \quad $^2$University of Toronto \quad $^3$Vector Institute \quad $^4$ETH Z\"{u}rich \\
}

\maketitle

In the supplementary material, we first provide details on our model design choices~(Sec.~\ref{sec:design_choices}). Then we describe the training details~(Sec.~\ref{sec:training_details}). Finally, we provide experiment details and additional results (Sec.~\ref{sec:experiment_details}). Please refer to the accompanied video for qualitative results on relighting and virtual object insertion.

\section{Model Details} 
\label{sec:design_choices}

\paragraph{Geometry definition.} 

Our method relies on an explicit surface definition for mesh extraction and efficient ray tracing. To this end, we follow NeuS~\cite{wang2021neus} and model the geometry with a Signed Distance (SD) Field  whose zero-level set defines the scene surface.
For volume rendering, the SDF values $f_\text{SDF}(\ray(t))$ are converted to opacity densities $\density(\ray(t))$ as:
\begin{equation}
\density(\ray(t)) = \max(\frac{-\frac{\text{d}\Phi_{\kappa}}{\text{d}t}(f_\text{SDF}(\ray(t)))}{\Phi_{\kappa}(f_\text{SDF}(\ray(t)))},0) 
\end{equation}
where $\Phi_{\kappa}(x) = \text{Sigmoid}(\kappa x) = \frac{1}{1+e^{-\kappa x}}$. Intuitively, the conversion is approximated by placing a unimodal function around the zero-level set of the SD field, \ie the derivative of the sigmoid function $\Phi^{'}_{\kappa}(x)$.
Here, $\kappa$ is a learnable parameter that controls the sharpness of the function and empirically $1/\kappa$ converges to zero as the training proceeds~\cite{wang2021neus}. To extract the mesh, we run marching cubes by querying the SD field on a predefined grid.

\vspace{-3mm}
\paragraph{Material definition.} We define the material properties of the scene using the physically-based (PBR) material model from Disney~\cite{Burley2012PhysicallyBasedSA}, which is a standard BRDF model adopted by modern graphics engines such as Unreal Engine~\cite{karis2013real}. 

The PBR material model represents the material properties using a 3-channel base color $\mathbf{k}_d \in \mathbb{R}^3$, and 2-channel specular properties $\mathbf{k}_s \in \mathbb{R}^2$. Here, $\mathbf{k}_s$ includes the roughness and metallic parameters. 
The metallic parameter is a real value $\in [0,1]$ indicating whether the surface behaves as a metal or nonmetal surface (e.g., plastic). 
Similarly, the roughness parameter is also a real value $\in [0,1]$ and defines how rough or smooth the surface is, thereby controlling how sharp or blurry reflections appear on that surface. 

In Fig.~\paperref{1}, \paperref{2}~and~\paperref{5} of the main paper, we visualize linear base color as an RGB image, and follow the graphics convention to visualize the specular properties $\mathbf{k}_s$ as a packed RGB image, where metallic is visualized with R-channel and roughness with G-channel.

\vspace{-3mm}
\paragraph{Normal extraction.} 
Recent works have proposed different ways to extract normal vectors from a neural field. For example, IRON~\cite{iron-2022} directly used the gradient of the underlying SD field, NeROIC~\cite{kuang2021neroic} used volume convolution, and Ref-NeRF~\cite{verbin2022refnerf} introduced an MLP network that predicts a normal vector at each point to regularize the noisy gradients of their volume density field.

In our method, we use an MLP $f_\text{norm.}$ to predict the normal direction for any 3D location, and estimate the normal vectors through volume rendering of the normal field. We further regularize the predicted normal directions to be consistent with normals computed from the gradient of SD Field. This design choice allows us to 
softly enforce the consistency with the underlying SDF geometry, while still maintaining the flexibility to account for high-frequency shading details with an MLP predicted normal (similar to a normal bump map in mesh-based representation~\cite{hasselgren2022nvdiffrecmc}).

\begin{figure*}[t]
\vspace{-1mm}
\centering
\includegraphics[width=1.0\linewidth,trim=0 0 0 0,clip]{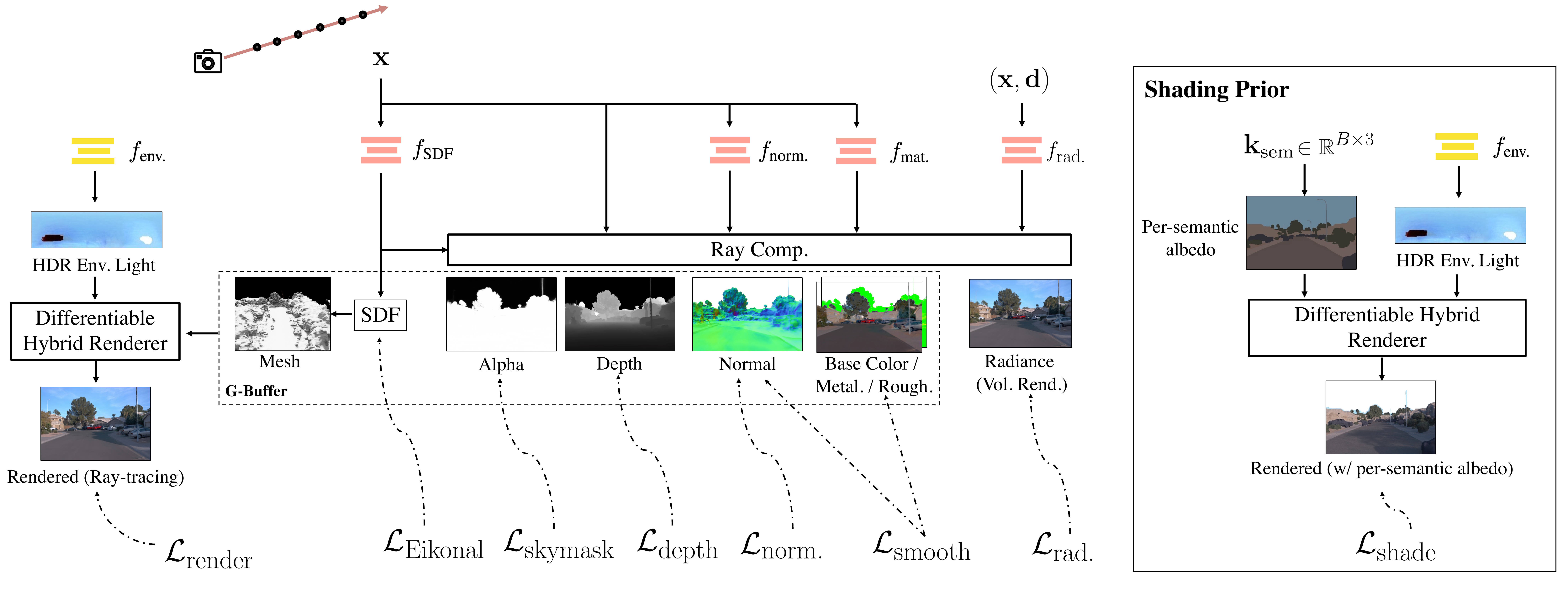}
\caption{\textbf{Training details for FEGR}. 
Similar to prior optimization-based inverse rendering methods~\cite{barron2014shape,boss2021nerd,hasselgren2022nvdiffrecmc}, FEGR adopts the major supervision of image reconstruction loss $\mathcal{L}_{\text{render}}$ and a set of regularization terms for each intrinsic property. 
}
\label{fig:supp_loss}
\end{figure*}

\vspace{-3mm}
\paragraph{Exposure and HDR to LDR conversion.} 
Real-world cameras often perform automatic white balancing and exposure correction~\cite{rematas2022urf} that result in inconsistent supervision signals across the input images.
To alleviate this issue, we additionally optimize a per-image exposure. 
Specifically, we optimize a set of variables $\{\beta_{i} \}_{i=1}^N$, where $\beta_i \in \mathbb{R}^3$ corresponds to the $i-$th image exposure compensation. During optimization, we normalize $\beta_i$ over all the images to resolve scale ambiguity: $\beta_{i} = \frac{\tilde{\beta}_{i}}{\frac{1}{N}\sum_{i=1}^N \tilde{\beta}_{i}}$. After that, we multiply the exposure compensation ratio $\beta_{i}$ with the predicted RGB values in the HDR linear RGB space.
During inference, we set all $\beta_{i}$ to a vector of ones.
Note that our lighting intensity and the rendering output of each pixels are HDR values in linear RGB space, while the ground truth values of each pixel in the captured image are in LDR sRGB space. To convert the predicted HDR RGB values to LDR sRGB, we use a standard gamma correction (gamma value equals to 2.2) and intensity clipping~\cite{li2020inverse,wang2021learning}.

\vspace{-3mm}
\paragraph{Implementation details.} 
The networks $f_\text{SDF}$, $f_\text{norm.}$ and $f_\text{mat.}$ are 2-layer MLPs with a multi-resolution hash positional encoding~\cite{mueller2022instant}, representing SDF, surface normal and material properties respectively. The dimension of the hidden layer is 64. 
The network $f_\text{env.}$ is 4-layer MLP with frequency positional encoding~\cite{mildenhall2020nerf} and exponential activation, representing HDR environment lighting. The hidden layer dimension is 256. 
For each primary ray, we sample 512 uniformly-spaced points and 64 adaptively sampled points following the scheme of NeuS~\cite{wang2021neus}. We sample 512 secondary rays via importance sampling over the BRDF and the HDR environment map. 
We extract the mesh using marching cubes~\cite{Lorensen87marchingcubes} with a $512\times 512\times 64$ grid, implemented in PyTorch~\cite{paszke2019pytorch} with CUDA support. 
Adapted from Nvdiffrecmc~\cite{hasselgren2022nvdiffrecmc}, the differentiable shading module is implemented in CUDA with OptiX~\cite{parker2010optix}. 
In the backward pass, we stop the gradient back-propagation to the extracted triangle meshes due to the GPU memory constraints. 
The inference time for marching cubes mesh extraction is 130ms. After each mesh update, the time to rebuild the bounding volume hierarchy (BVH)~\cite{parker2010optix} is 75ms. Note that the mesh extraction and BVH are only computed once per scene during inference. The shading pass of one 640x960 image takes 210ms.

\section{Training Details} 
\label{sec:training_details} 

\begin{figure*}[t!]
\vspace{-1mm}
\centering
\begingroup
\setlength{\tabcolsep}{2pt}
\includegraphics[width=1\linewidth]{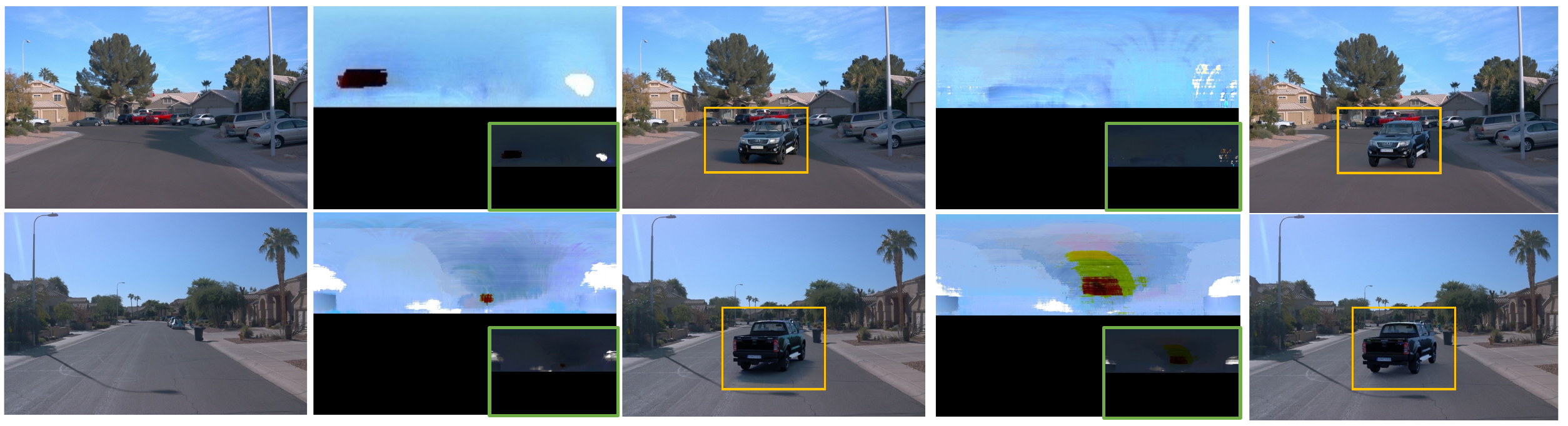}\vspace{-2mm} 
\resizebox{0.99\textwidth}{!}{
\begin{tabular}{C{0.2\textwidth}C{0.2\textwidth}C{0.2\textwidth}C{0.2\textwidth}C{0.2\textwidth}}
Training image & \multicolumn{2}{C{0.4\textwidth}}{Ours}  & \multicolumn{2}{C{0.4\textwidth}}{Ours (w/o $\mathcal{L}_\text{shade}$)} \\
\end{tabular}
}
\endgroup
\vspace{-3mm}
\caption{\textbf{Qualitative ablation of shading prior.} 
We qualitatively ablate the effect of the semantic-aware shading regularization loss $\mathcal{L}_\text{shade}$. For each scene, we visualize the estimated HDR environment map and an object insertion result. On the bottom-right of the environment map, we divide the HDR value by 30 to better display the HDR component of the environment map. 
}
\label{fig:supp_ablate_shading_prior} 
\end{figure*}

In the following, we provide additional details for each loss function, as well as an intuitive explanation of their contribution to the combined optimization.
An overview of our training pipeline is provided in Fig.~\ref{fig:supp_loss}. Except for shading prior loss $\mathcal{L}_{\text{shade}}$, similar loss terms were used before in the literature. The ablation study provided in Sec.~\ref{sec:experiment_details} and Fig~\ref{fig:supp_ablate_shading_prior} therefore focuses on the $\mathcal{L}_{\text{shade}}$.

\vspace{-3mm}
\paragraph{Shading prior $\mathcal{L}_{\text{shade}}$.} 
As is described in the main paper in Sec.~\paperref{3.3}, the motivation for introducing the semantics-aware shading regularization term $\mathcal{L}_{\text{shade}}$ is to regularize the \textit{lighting}. Indeed, $\mathcal{L}_{\text{shade}}$ encourages that the shadows present in the input images are explained by the combination of lighting and geometry, instead of degenerating into an easy solution of baking them into albedo. 

To this end, we introduce an auxiliary piecewise-constant albedo representation and encourage its re-rendering to be consistent with the groundtruth image. Intuitively, due to the limited capacity of the piecewise-constant albedo representation, the supervision signal emerging from the lighting effects will be mainly propagated to the HDR environment light. Specifically, we initialize each semantic class with a 3-channel albedo value, which we optimize during training. Thereby semantic segmentation labels are computed with an off-the-shelf semantic segmentation network~\cite{tao2020hierarchical}. To compute the $\mathcal{L}_{\text{shade}}$, we use the estimated lighting to render this per-semantic class albedo and encourage the rendered result to be consistent with the groundtruth images. 

We depict an example of the per-semantic class albedo in Fig.~\ref{fig:supp_loss} (right). 
In practice, we apply this loss on the semantic classes \textit{road, sidewalk, building, wall} which typically have a single dominant albedo and provide informative visual cues such as boundary of cast shadows. We ablate the effect of this loss in Sec.~\ref{sec:experiment_details} and Fig~\ref{fig:supp_ablate_shading_prior}.

\vspace{-3mm}
\paragraph{Regularization terms} $\mathcal{L}_{\text{reg.}}$ denotes the weighted sum of additional regularization terms: $\mathcal{L}_{\text{smooth}}$, $\mathcal{L}_{\text{Eikonal}}$, $\mathcal{L}_{\text{skymask}}$. 

We follow prior works~\cite{wang2021neus,yariv2021volume} and regularize the gradient of SDF value $\sdf$ with an Eikonal term:
\begin{equation}
  \mathcal{L}_{\text{Eikonal}} = \frac{1}{|\mathcal{X}|} \sum_{\mathbf{x} \in \mathcal{X}} (||\nabla_\mathbf{x}\sdf(\mathbf{x})||_2 - 1)^2,
\end{equation}
where $\mathcal{X}$ is the set of points sampled along the ray. 

Similar to prior inverse rendering works~\cite{land1971lightness1,barron2014shape,munkberg2021nvdiffrec,hasselgren2022nvdiffrecmc}, we also encourage local smoothness of normals and material properties. 
Specifically, we follow Nvdiffrecmc~\cite{hasselgren2022nvdiffrecmc} and apply the smoothness regularization for base color $\mathbf{k}_d$, normal $\mathbf{n}$, and material $\mathbf{k}_s$:
\begin{align}
  \mathcal{L}_{\text{smooth}} = &
  \frac{1}{|\mathcal{X}|} \sum_{\mathbf{x} \in \mathcal{X}} |\mathbf{k}_d(\mathbf{x}) - \mathbf{k}_d(\mathbf{x} + \mathbf{\epsilon})| \nonumber\\
  &+ \frac{1}{|\mathcal{X}|} \sum_{\mathbf{x} \in \mathcal{X}} |\mathbf{k}_s(\mathbf{x}) - \mathbf{k}_s(\mathbf{x} + \mathbf{\epsilon})| \nonumber\\
  &+ \frac{1}{|\mathcal{X}|} \sum_{\mathbf{x} \in \mathcal{X}} |\mathbf{n}(\mathbf{x}) - \mathbf{n}(\mathbf{x} + \mathbf{\epsilon})|, 
\end{align}
where $\mathbf{\epsilon} \sim \mathcal{N}(0, \sigma=0.02)$ is a local perturbation vector. 

Finally, prior works such as NeRF-OSR~\cite{rudnev2022nerfosr} do not explicitly handle the sky and hence produce many floaters in their scene representation. In our work, we follow~\cite{wang2022neural} and apply a binary cross entropy (BCE) loss $\mathcal{L}_{\text{skymask}}$ between the volume rendered alpha channel and the sky semantic segmentation masks. 
The sky masks are again obtained from an off-the-shelf semantic segmentation network~\cite{tao2020hierarchical}. 
In practice, we assign a small weight to this sky mask regularization to only carve out the floaters in the sky region, while not harming the geometry of the scene.

\vspace{-3mm}
\paragraph{Training details.} The final loss is a weighted sum of the reconstruction and regularization terms
\begin{align}
\mathcal{L} =& \mathcal{L}_{\text{render}} + \lambda_\text{depth}\mathcal{L}_{\text{depth}} + \lambda_\text{rad.}\mathcal{L}_{\text{rad.}} \nonumber \\
& + \lambda_\text{norm.}\mathcal{L}_{\text{norm.}} + \lambda_\text{shade}\mathcal{L}_{\text{shade}} \nonumber\\
&+ \lambda_\text{Eikonal}\mathcal{L}_{\text{Eikonal}} + \lambda_\text{skymask}\mathcal{L}_{\text{skymask}} \nonumber \\
&+ \lambda_\text{smooth}\mathcal{L}_{\text{smooth}}. 
\end{align}
where the weight of each loss function is set to: $\lambda_\text{rad.}=\lambda_\text{norm.}=1$, $\lambda_\text{shade}=0.1$, $\lambda_\text{Eikonal}=0.05$, $\lambda_\text{skymask}=\lambda_\text{smooth}=0.01$. $\lambda_\text{depth}$ is set to 1 on Driving data and 0 for NeRF-OSR dataset~\cite{rudnev2022nerfosr}.
We use Adam optimizer~\cite{kingma2017adam} with a learning rate of 1e-2. 
As the mesh extraction requires a well initialized SD field, we run a warm-up phase for 5k iterations in which we remove  $\mathcal{L}_{\text{render}}$ and $ \mathcal{L}_{\text{shade}}$. After the warm-up phase we continue optimizing all the loss terms for additional 50k iterations. 
In each batch, we sample 4096 rays. With the parameters detailed above, {\ourmodel} consumes about 20GB GPU memory during training.

\begin{figure*}[t!]
\vspace{-1mm}
\centering
\begingroup
\setlength{\tabcolsep}{2pt}
\includegraphics[width=1\linewidth]{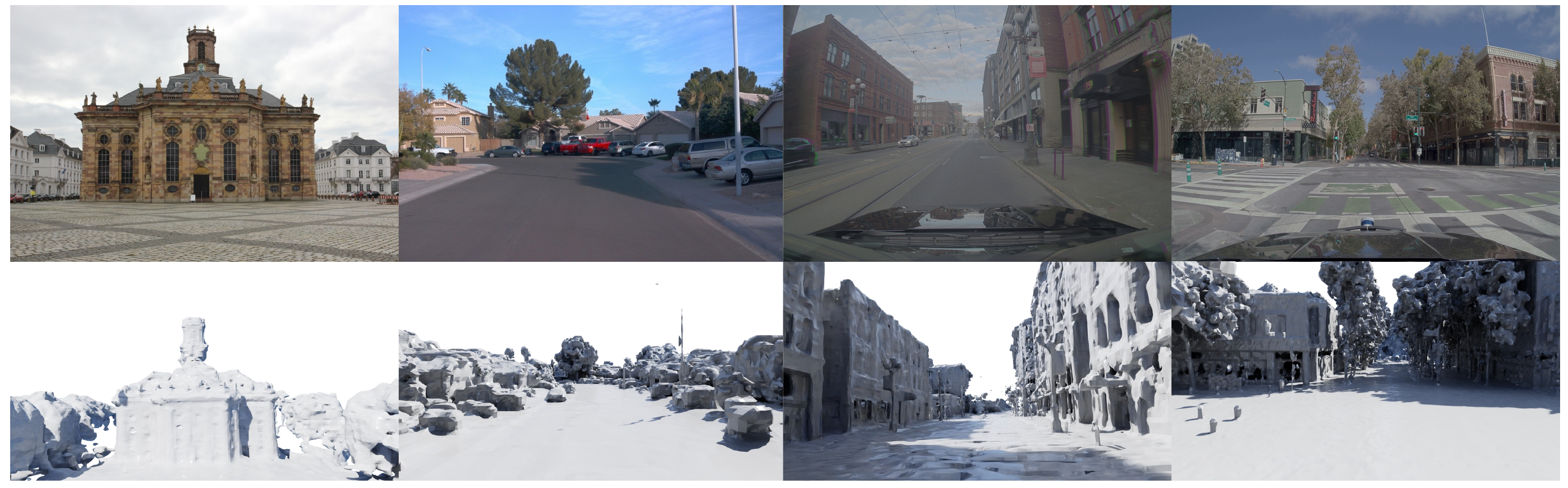}\vspace{-2mm} 
\endgroup
\caption{Qualitative visualization of mesh reconstruction. 
We visualize the underlying geometry reconstructed by our method. 
}
\label{fig:supp_mesh} 
\end{figure*}

\section{Experiment Analysis and Results}
\label{sec:experiment_details} 
In this section, we provide a detailed experimental setup and additional results. 

\vspace{-3mm}
\paragraph{Relighting details.} The application of \textit{relighting} aims to generate imagery of the 3D scene under the lighting conditions specified by the users, typically an HDR environment map. 
FEGR represents the scene with standard PBR materials, and thus can directly replace the reconstructed HDR environment light $f_\text{env.}$ with the user-specified lighting. 

The NeRF-OSR~\cite{rudnev2022nerfosr} baseline requires spherical harmonics lighting, and thus we converted the HDR map to an SH representation as suggested in the paper\footnote{We use this~\href{https://github.com/chalmersgit/SphericalHarmonics}{repository} to estimate the SH coefficients}. 
In the qualitative comparison (main paper and the accompanied video), we tackle a more challenging scenario and use a high-contrast HDR map with strong directional light to highlight the ability of the methods to cast shadows. 
In this case, NeRF-OSR shows relatively worse qualitative performance, with reasons in twofold:
(i) The strong directional sunlight makes the small normal artifacts more pronounced, and (ii) The SH coefficients estimated from peaky HDR environment maps are not on the training data manifold, making the shadow network fail to generalize. 
In addition, NeRF-OSR implicitly represents shadows with an MLP learned across multiple illumination, and thus cannot guarantee that the shadows follow the rule of light transport.

Compared to NeRF-OSR, FEGR supports rendering the physics-based shadow effects from the user-specified lighting via ray-tracing, such as shadows due to self-occlusion. 
We refer to the accompanied video for qualitative comparison and additional results on relighting.

\vspace{-3mm}
\paragraph{Object insertion details.} 
The application of \textit{virtual object insertion} takes as input synthetic objects with know geometry and materials, and aims to produce photorealistic imagery by placing them into real-world images. 
This requires proper handling of lighting effects such as cast shadows and specular highlights.
For this image editing task, we follow the object insertion formulation in~\cite{wang2022neural}, which first separately renders the foreground objects and scene shadows, and then composite them onto the input scene image. The rendering is performed in Blender~\cite{blender}. 

Existing works on inverse rendering~\cite{bi2020deep,boss2021nerd,physg2021,zhang2021nerfactor,iron-2022,zhang2022invrender,munkberg2021nvdiffrec,hasselgren2022nvdiffrecmc,rudnev2022nerfosr} typically adopt simplified lighting representations such as a point light~\cite{nerv2021,bi2020neural} or low-frequency spherical lobes~\cite{boss2021nerd,rudnev2022nerfosr}. These works do not aim to estimate spatially-varying lighting. Instead, they only use lighting as a side-product in the joint optimization process and they discarded it after training. 

We compare FEGR on the task of virtual object insertion with recent state-of-the-art learning-based outdoor lighting estimation methods~\cite{hold2019deep,wang2022neural}. Qualitative comparison is available in main paper Fig.~\paperref{6} and a user study in main paper Table~\paperref{3}. 
For the user study, we follow the setup of~\cite{wang2022neural} and conduct it on Amazon Mechanical Turk. 
Compared to learning-based feed-forward lighting estimation models, we acknowledge that our method consumes more information as input and requires online optimization. However, we stress that our method achieves significantly improved results and recovers accurate shadow direction and intensity, which is challenging for single-image feed-forward methods. We believe that our formulation can inspire future works on the role of lighting in optimization-based inverse rendering. 

We refer to the accompanied video for additional results on virtual object insertion. 

\vspace{-3mm}
\paragraph{Qualitative ablation of shading prior $\mathcal{L}_\text{shade}$.} 
We qualitatively ablate and show the results in Fig.~\ref{fig:supp_ablate_shading_prior}. 
When training without the shading prior loss term $\mathcal{L}_\text{shade}$, the estimated environment light can still predict the peak direction but typically fails to produce sharp cast shadows and correct shadow scale. 
This indicates the shading prior $\mathcal{L}_\text{shade}$ is beneficial for HDR light estimation. 

\vspace{-3mm}
\paragraph{Qualitative visualization of meshes.} 
In Fig.~\ref{fig:supp_mesh}, we visualize the underlying geometry extracted by marching cubes. In the hybrid rendering described in main paper Sec.~\paperref{3.2}, the mesh accounts for the visibility query of secondary rays to render cast shadows.

{\small
\bibliographystyle{ieee_fullname}
\bibliography{egbib}
}